\documentclass{article}

\usepackage{PRIMEarxiv}

\usepackage[utf8]{inputenc} 
\usepackage[T1]{fontenc}    
\usepackage{hyperref}       
\usepackage{url}            
\usepackage{booktabs}       
\usepackage{amsfonts}       
\usepackage{nicefrac}       
\usepackage{microtype}      
\usepackage{lipsum}
\usepackage{fancyhdr}       
\usepackage{graphicx}       
\graphicspath{{media/}}     
\usepackage{graphicx}
\usepackage{booktabs}
\usepackage{multirow}
\usepackage{wrapfig}
\usepackage{floatrow}

\newfloatcommand{figurebox}{figure}[\nocapbeside][\dimexpr(\textwidth-\columnsep)/2\relax]
\newfloatcommand{tablebox}{table}[\nocapbeside][\dimexpr(\textwidth-\columnsep)/2\relax]

\usepackage{amssymb}
\usepackage{graphicx}
\usepackage{array}
\newcolumntype{C}[1]{>{\centering\arraybackslash}p{#1}}
\newcolumntype{L}[1]{>{\arraybackslash}p{#1}}
\newcolumntype{R}[1]{>{\raggedleft\arraybackslash}p{#1}}

\usepackage{algorithm}
\usepackage{algorithmic}
\usepackage{amsmath}

\usepackage{tabularx} 
\usepackage{booktabs} 

\usepackage[accsupp]{axessibility}
\title{EMIE-MAP: Large-Scale Road Surface Reconstruction Based on Explicit Mesh and Implicit Encoding
}

\author{
  Wenhua~Wu \\
  Shanghai Jiao Tong University  \\
   \And
  Qi~Wang \\
  Shanghai Jiao Tong University \\
  \And
  Guangming~Wang \\
  University of Cambridge \\
  \And
  Junping~Wang \\
  Hozon New Energy Automobile Co., Ltd \\
  \And
  Tiankun~Zhao \\
  Hozon New Energy Automobile Co., Ltd \\
  \And
  Yang~Liu \\
  Hozon New Energy Automobile Co., Ltd \\
  \And
  Dongchao~Gao \\
  Hozon New Energy Automobile Co., Ltd \\
  \And
   Zhe~Liu \\
   Shanghai Jiao Tong University \\
  \And
  Hesheng~Wang \thanks{Corresponding Author.}\\
  Shanghai Jiao Tong University \\
}

\begin{document}
\maketitle
\begin{abstract}
Road surface reconstruction plays a vital role in autonomous driving systems, enabling road lane perception and high-precision mapping. Recently, neural implicit encoding has achieved remarkable results in scene representation, particularly in the realistic rendering of scene textures. However, it faces challenges in directly representing geometric information for large-scale scenes. To address this, we propose EMIE-MAP, a novel method for large-scale road surface reconstruction based on explicit mesh and implicit encoding. The road geometry is represented using explicit mesh, where each vertex stores implicit encoding representing the color and semantic information. To overcome the difficulty in optimizing road elevation, we introduce a trajectory-based elevation initialization and an elevation residual learning method based on Multi-Layer Perceptron (MLP). Additionally, by employing implicit encoding and multi-camera color MLPs decoding, we achieve separate modeling of scene physical properties and camera characteristics, allowing surround-view reconstruction compatible with different camera models. Our method achieves remarkable road surface reconstruction performance in a variety of real-world challenging scenarios.


\end{abstract}    
\section{Introduction}
    
\label{sec:intro}
\begin{figure*}[t] 
\center{\includegraphics[width=1.0\textwidth]{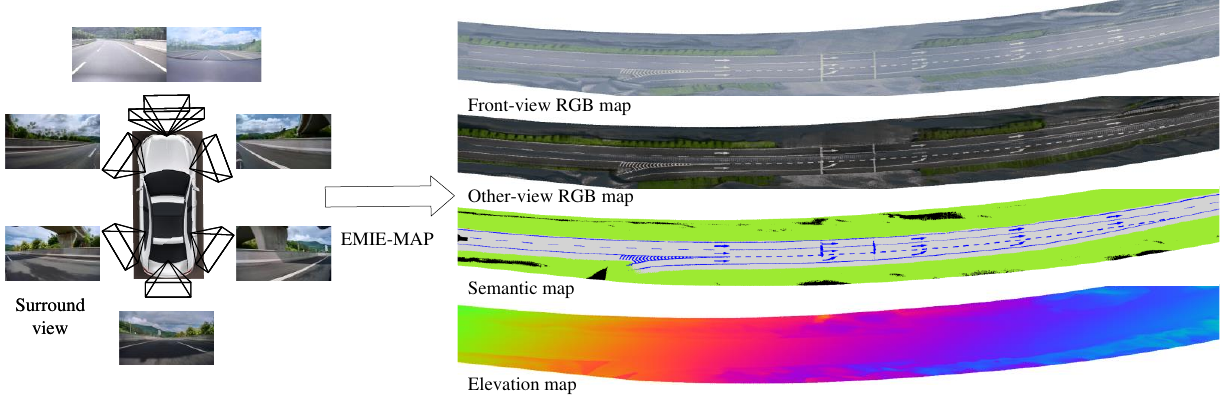}} 
\caption{ We propose EMIE-MAP, a novel large-scale road surface reconstruction method based on \textbf{E}xplicit \textbf{M}esh and \textbf{I}mplicit \textbf{E}ncoding. By taking input surround-view video sequences and localization information, EMIE-MAP is capable of reconstructing RGB maps, semantic maps, and elevation maps. The RGB maps are corresponding to different cameras. EMIE-MAP integrates the advantages of explicit and implicit representations, enabling accurate road surface reconstruction and rendering.
        }
\label{first}
\end{figure*}

The development of autonomous driving systems has brought about a growing need for accurate road reconstruction, as it plays an essential part in enabling lane perception and high-precision mapping of roads. In recent years, Bird’s Eye View (BEV) perception has gained prominence in the field of autonomous driving due to its natural alignment with downstream tasks such as planning and control. As a result, the importance of large-scale road reconstruction has been highlighted, especially its potential application in providing training and validation data for perception tasks of autonomous vehicles.

Historically, 3D reconstruction methods are divided into traditional methods~\cite{ozyecsil2017survey} and Neural Radiance Fields (NeRF)-based methods~\cite{mildenhall2021nerf}. Although the traditional multi-view stereo methods can effectively reconstruct dense or semi-dense points, they often produce incomplete or noisy results when applied to untextured road. NeRF-based approaches, on the other hand, show promise in photorealistic reconstruction, but are limited by high resource consumption, especially in large-scale scenarios.

The work most similar to this paper is RoMe~\cite{mei2023rome}. It is a simple and effective method for large-scale road reconstruction using grid representation. To address the complexity of the task, RoMe decomposes the 3D road into a triangular grid and implicitly models the road elevation using a multi-layer perception (MLP). Each mesh vertex includes additional properties of color and semantics to capture fine surface details. However, it is inaccurate to directly regress ground height from positional encoding. Luminosity inconsistencies between surround views can cause convergence instability of RoMe. To address these challenges, we propose a large-scale road reconstruction method (EMIE-MAP) based on explicit grid and implicit encoding.

EMIE-MAP utilizes a combination of explicit mesh and implicit encoding to improve accuracy and efficiency. An explicit mesh representation is used to capture the road geometry, and each mesh vertex stores an implicit encoding that encapsulates color and explicit semantic information. To meet the challenge of road elevation optimization, we propose a trajectory-based elevation initialization method supplemented by an elevation residual learning method based on MLP. This combination allows us to achieve robust and accurate road reconstruction, especially in the case of large changes in road elevation. In addition, our approach combines multi-camera RGB MLPs implicit decoding, enabling independent modeling of scene physical properties and camera characteristics. This facilitates surround-view reconstruction compatible with a wide range of camera types, enhancing the versatility and applicability of our approach in a variety of autonomous driving systems. By unwrapping the representation of scene attributes and camera characteristics, we are able to achieve a more comprehensive and adaptable large-scale road reconstruction framework to meet the specific requirements of different real-world scenarios and environments.

The main contributions of our work can be summarized as follows:

\begin{itemize}

  \item We propose EMIE-MAP, a large-scale road surface reconstruction method. The core of EMIE-MAP is a road surface representation based on explicit mesh and implicit encoding, facilitating the storage and optimization of road geometry, color, and semantic information.

  \item We introduce a trajectory-based elevation initialization and an MLP-based elevation residual prediction method to learn ground elevation, overcoming the difficulties in reconstructing road surfaces on slopes.

  \item We propose a color optimization method based on consistent implicit color encoding and multi-camera color decoding, enabling the separation of scene physical attributes and camera characteristics. This allows the method to be applied to surround-view with different camera models.

  \item Experiments conducted in open source datasets and various real challenging scenes demonstrate that EMIE-MAP exhibits remarkable reconstruction results.
\end{itemize}

\section{Related work}
\label{sec:related_work}

\subsection{Explicit 3D Reconstruction}

Explicit 3D reconstruction methods directly reconstruct 3D objects into point clouds or meshes. There has been a lot of accumulation of these algorithms~~\cite{ozyecsil2017survey}. Incremental methods of ~\cite{snavely2006photo, snavely2011scene, wu2013towards, moulon2013adaptive} are a series of classical Structure from Motion  (SfM) methods. COLMAP~\cite{Schonberger2016structure, schonberger2016pixelwise} extracts feature points from image data to reconstruct 3D scenes, but it does not perform well in weak texture areas such as road surface, and the reconstructed points are too sparse. Sameer Agarwal \emph{et al.}~\cite{agarwal2011building} use a large number of images on the Internet to complete the reconstruction of urban scale, but this method also faces the problems of sparse results and sensitivity to texture. Algorithms that directly target road reconstruction can produce dense outputs ~\cite{fan2018road, yu20073d, brunken2020road, guo2015automatic, fan2021rethinking}, but they are limited to the reconstruction of small road areas. Methods for representing a 3D scene in explicit meshes can automatically texture the model in large-scale 3D reconstruction~\cite{levoy2023light, waechter2014let}. They can improve the visual effect of the reconstructed model and provide a new direction for 3D reconstruction.

\subsection{Implicit 3D Reconstruction}
NeRF~\cite{mildenhall2021nerf} is one of the most fundamental works to pioneer implicit 3D reconstruction. It uses deep learning methods to learn an implicit scene representation from existing images from different perspectives, and can render the scene to generate a simulated picture from a new perspective. The utilization of additional point cloud inputs can improve the effectiveness of NeRF~\cite{rematas2022urban, li2023read}. There are a lot of works to improve NeRF or expand its use. Block-NeRF~\cite{tancik2022block} divides large scenes into blocks and trains the NeRF network separately. These networks are trained in parallel and joined together for inference. SUDS ~\cite{turki2023suds} applies NeRF to scalable urban dynamic scenarios. GM-NeRF ~\cite{chen2023gm} learns generalized model-based neural radiation fields from multi-view images. Andreas Meuleman \emph{et al.} ~\cite{meuleman2023progressively} propose asymptotically optimized local radiation fields for robust view synthesis. ABLE-NeRF ~\cite{tang2023able} proposes a volume framework based on self-attention mechanism and introduces learnable embedded features to capture perspective-dependent effects in scenes. SPARF ~\cite{truong2023sparf} solves the problem of sparse view and inaccurate pose in neural radiation field. Recently, there are multiple works using diffusion reconstruct an implicit 3D surface in high fidelity from a cloud of noise points ~\cite{zhou2023sparsefusion, anciukevivcius2023renderdiffusion, wang2023alto, melas2023pc2}.

Explicit 3D reconstruction can directly represent 3D scenes without rendering operations, and implicit representation schemes such as NeRF can model rich texture details. Combining the advantages of the above two schemes, RoMe~\cite{mei2023rome} models the color and semantic information of the road surface into explicit meshes. For the road elevation information, it is stored in the implicit MLP structure. RoMe uses a purely visual approach to reconstruct large areas of road using sampled surround-view camera images. Each camera goes through the same process to produce the final scene, which makes it impossible to determine whether the scene is optimized in a brighter or darker direction when each camera's luminosity is inconsistent. There are some methods ~\cite{schonberger2016pixelwise, waechter2014let, rematas2022urban} proposing solutions for luminosity inconsistency, but they focus on the luminosity changes of the same camera at different times. Those methods ignore the inherent luminosity differences among various surround-view cameras carried by the autonomous vehicle. On the other hand, vehicle trajectory information can provide good prior for ground elevation prediction, and using MLP to predict elevation residual results can get more accurate elevation results. EMIE-MAP improves on both.

\begin{figure*}[t] 
\center{\includegraphics[width=1.0\textwidth]{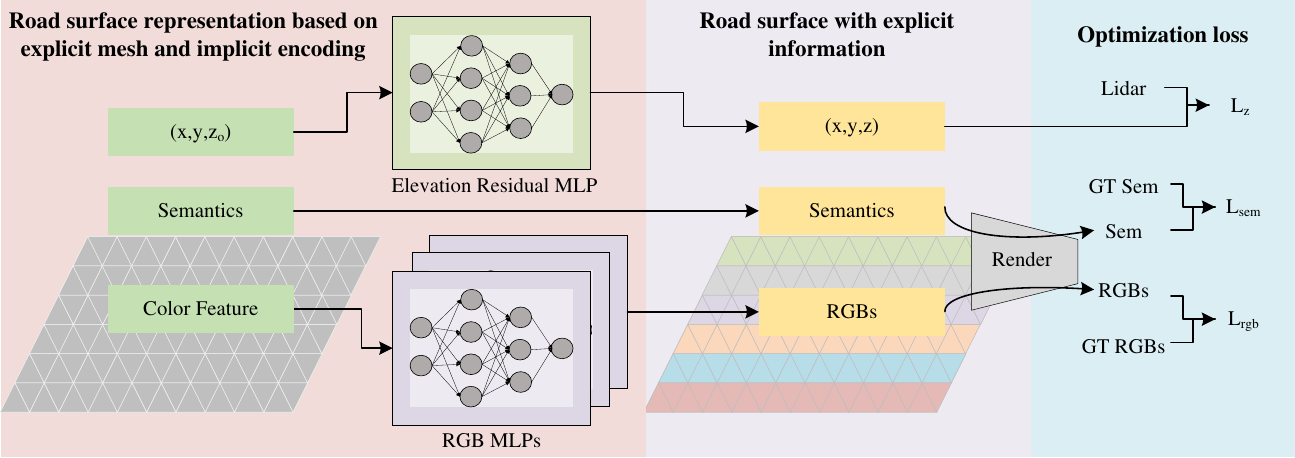}} 
\caption{Overview of EMIE-MAP. The left side presents the proposed road surface representation based on explicit mesh and implicit encoding. We utilize a mesh composed of equilateral triangular faces to represent the road structure. Each vertex stores its initial coordinates $(x,y,z_0)$, semantic information, and implicit color encoding. An elevation residual network predicts the elevation residual at each vertex, while multiple RGB MLPs decode the implicit color features into observed colors for the corresponding camera. This yields a road surface with explicit information in the framework's middle section. The right side shows optimization losses. The generated RGB and semantic maps through direct rendering are supervised by observed images. Additionally, Lidar point clouds are utilized to supervise the road surface elevation.}
\label{framework}
\end{figure*}

\section{Method}
\label{sec:method}
The overview of the proposed EMIE-MAP is shown in Fig.~\ref{framework}. The left side presents a road surface representation based on explicit mesh and implicit encoding. We utilize a mesh composed of equilateral triangular faces to represent the road structure. Each vertex stores its initial coordinates $(x,y,z_0)$, semantic information, and implicit color encoding. An elevation residual MLP predicts the elevation residual, while multiple RGB MLPs decode the implicit color features into observed colors for the corresponding camera. This yields a road surface with explicit information in the middle section of the framework. On the right side, optimization losses are employed. Through direct rendering, generated RGB and semantic maps are supervised by observed images. Additionally, Light Detection and Ranging (Lidar) point clouds are utilized to supervise the road surface elevation. In this section, we will provide a detailed description of EMIE-MAP.

\subsection{Road Surface Representation based on Explicit mesh and Implicit Encoding}
Traditional reconstruction algorithms use point clouds, meshes, or voxels for scene representation, directly displaying scene information. However, they face challenges such as holes, difficult optimization, and a lack of flexibility in representation. In contrast, NeRF~\cite{mildenhall2021nerf} provide a purely implicit scene representation that uses MLP to model the mapping from spatial coordinates to spatial geometry and color information, resulting in a more flexible scene representation. However, on one hand, NeRF~\cite{mildenhall2021nerf} requires substantial computational resources for image rendering through ray sampling integration. On the other hand, while the rendering results are realistic, it is challenging to obtain the explicit geometric information of the scene.

To address these issues, we propose a road surface representation method that combines an explicit mesh and implicit encoding. Specifically, we use a mesh composed of equilateral triangular faces to represent the scene structure. The length of each face is $a$. Each vertex stores two explicit attributes: position $(x,y,z_0)$ and semantics $sem$, along with an implicit color encoding $l_c$. Additionally, we employ a MLP to estimate elevation residuals, obtaining more accurate position information. Multiple RGB MLPs decode the implicit color features into observed colors for the corresponding camera. Next, we will look in detail at the three road surface attributes of elevation, color, and semantics.

\noindent \textbf{Elevation.} RoMe~\cite{mei2023rome} directly uses an MLP to predict elevations, which is effective for relatively flat road surfaces. However, when the road surface has steep slopes, simple MLPs struggle to predict drastic elevation changes. To address the difficulty, we propose an elevation prediction approach based on trajectory elevation initialization and residual prediction. Unlike RoMe ~\cite{mei2023rome}, we no longer initialize the road surface as a horizontal plane. Based on the fact that the vehicle runs close to the ground along the lane, we use trajectory elevation to initialize the road surface elevation. Then, we employ an MLP to predict elevation residuals. Instead of directly predicting drastic elevation changes, the MLP only needs to predict low-frequency elevation residuals that are easier to learn. Specifically, the coordinate $(x,y)$ is encoded and input to the residual prediction network $MLP_{hr}$. The final road surface elevation is the sum of the initial elevation and the residual:
\begin{equation}
z_r = MLP_{hr}(PE(x,y)),\\
z_f = z_0+z_r,
\end{equation}
where $PE()$ refers to Positional Encoding, which employs a combination of sine and cosine functions to generate positional encoding vectors.

\noindent \textbf{Color.} In order to accommodate different perceptual needs, the surround-view cameras equipped in automobiles vary. Generally, the front view, serving as the primary perspective, utilizes cameras with a wider field of view and higher resolution. Different camera models result in variations in the RGB images when observing the same area, which are manifested in terms of brightness and saturation. In such cases, directly optimizing the explicit RGB values for RoMe ~\cite{mei2023rome} strategy leads to optimization conflicts. To address this challenge, we employ implicit color encoding $l_c$ to replace the explicit RGB values. For different camera models, we utilize different decoders for color decoding:
\begin{equation}
rgb_i = MLP_{rgb\_i}(l_c).
\end{equation}

The decoded colors are supervised against the observed colors from their respective cameras. It should be noted that different camera decoders share the same implicit color encoding. Implicit color encoding represents the physical attributes of the scene, while the decoders learn camera characteristics, thus achieving modeling of inconsistent observations from multiple cameras and ultimately obtaining consistent scene information.

\noindent \textbf{Semantics.} Unlike color, semantics are intrinsic properties of the scene and independent of camera. Therefore, we directly utilize explicit parameters to represent semantics.

\begin{figure*}[t] 
\center{\includegraphics[width=\textwidth]{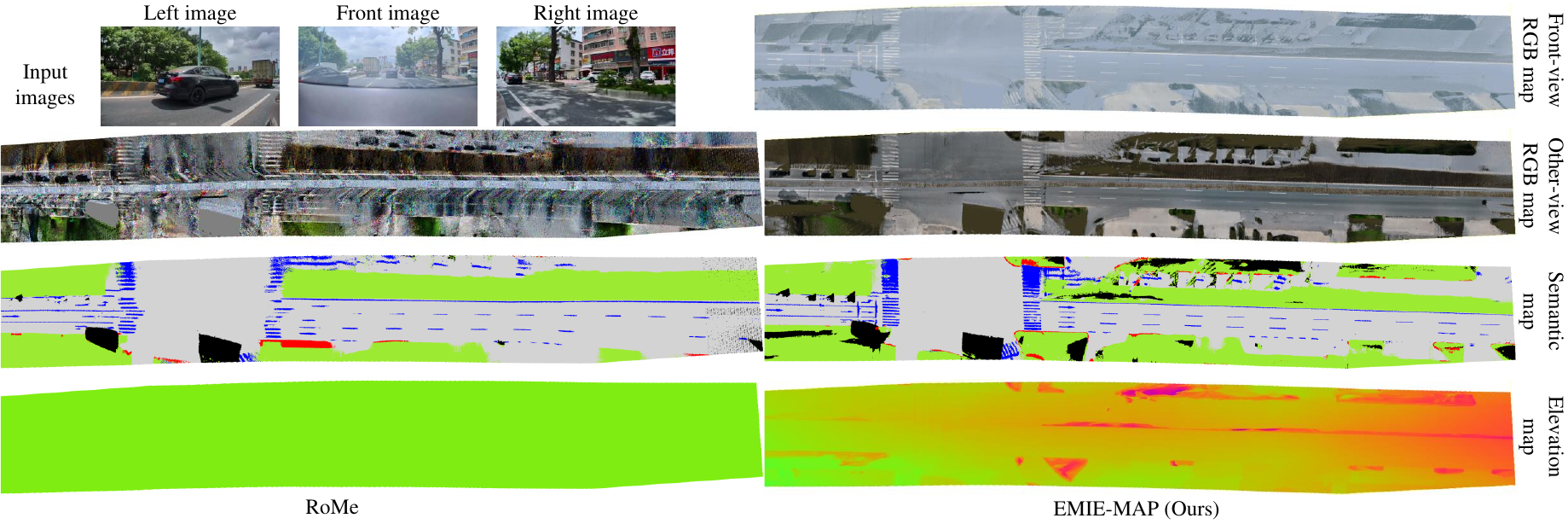}} 
\caption{Road surface reconstruction results in city street scene. From top to bottom are the input images, reconstructed road surface RGB images from different camera perspectives, semantic map, and elevation map. 
Despite the presence of dense traffic, the road surface in occluded areas is effectively filled in by images taken at different times.}
\label{city}
\end{figure*}

\subsection{Optimizing Strategies}
\noindent \textbf{Mesh initialization.} By utilizing localization algorithm based on multiple sensors such as Inertial Measurement Unit (IMU), wheel encoders, Global Positioning System (GPS), and cameras, we can acquire high-precision vehicle trajectory information in both the horizontal and vertical directions. Under normal circumstances, the vehicle adheres to the road surface and travels along the lane. Therefore, we can extend the trajectory in the horizontal direction to obtain the road area. The extension distance is determined by the width of the road. After determining the road surface area, we construct a road surface model based on a designed representation. The elevation of the road $z$ is initialized through interpolation of nearby trajectory points' elevations $z_0$. Semantics $sem$ and implicit color encoding $l_c$ are initialized randomly. 

\noindent \textbf{Data sampling and observation area sampling.} 
To accelerate the training process, we sample a batch of $B$ images for each training iteration. Since the observed road surface area varies at different positions, we employ a trajectory-based data sampling strategy to improve computational efficiency. The images within each batch are obtained from observations at adjacent trajectory points, resulting in these images having a similar observation area. Subsequently, we extract the road surface within an $80 m$ distance before and after the trajectory point as the observation area for the batch of data. By employing this data sampling and observation area sampling strategy, each training iteration only requires explicit processing, rendering, and other operations on a small segment of the road surface, significantly speeding up the reconstruction process.

\noindent \textbf{Road surface visualization.} After determining the observation area for each training batch, we need to fully visualize the partially observed road surface. Specifically, a elevation residual network is used to predict the elevation residual at each position, which is then added to the initialized elevation to obtain the road surface elevation information. The implicit color encoding is decoded using the corresponding color decoder for each camera, resulting in explicit RGB values for visualization. The semantics are directly used from the stored semantic information. This process ultimately yields a fully visualized road surface, including coordinate, color, and semantic information.

\noindent \textbf{Rendering.} Unlike NeRF~\cite{mildenhall2021nerf} that samples along the observation rays and integrates information at the sampled points, we employ a direct projection rendering method based on the pinhole camera principle. Given the road surface mesh, camera poses $T$, and camera intrinsic parameters $K$, the pixel coordinates $(u,v,1)$ corresponding to each road surface point $(x,y,z)$ can be obtained using the following formula:
\begin{equation}
(u,v,1)^T = K\cdot T \cdot(x,y,z,1)^T.
\end{equation}

\begin{figure*}[t] 
\center{\includegraphics[width=\textwidth]{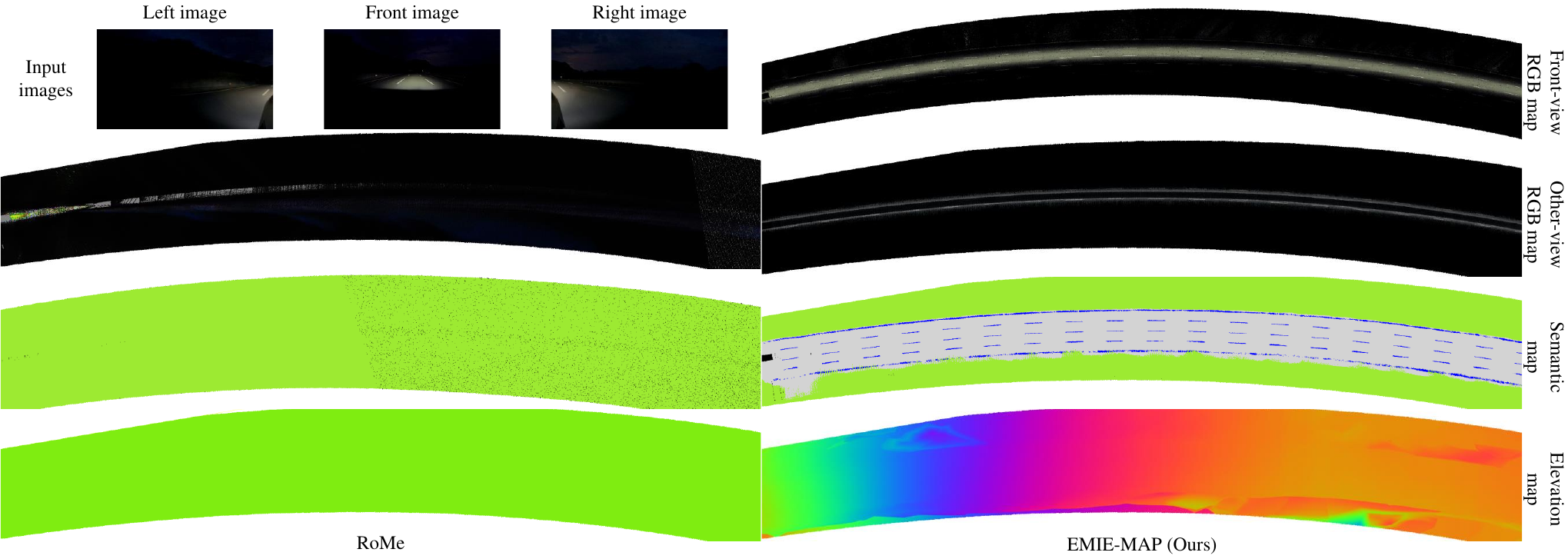}} 
\caption{Road surface reconstruction results in night scene. The only light source is the car headlights. In the input image with poor lighting quality, the RGB results are also dark. However, our method is still able to reconstruct the desired semantic map.}
\label{night}
\end{figure*}

The color and semantic information of the ground points are then projected onto the corresponding pixel, resulting in the rendered RGB image $C$ and semantic image $S$. It is important to note that for the rendering of RGB images from each camera, we utilize the corresponding decoded RGB values. Compared to volume rendering, our rendering approach better simulates the process of acquiring real images and significantly reduces computational complexity. In reality, the rays emitted from the surface of an object, pass through the camera center and reach the pixel plane. Our rendering process simulates the process of capturing real images.

\noindent \textbf{Training loss}
For color, we supervise the rendering of RGB images by using observed images from the cameras. For semantics, we utilize a pre-trained Mask2Former~\cite{cheng2022masked} for semantic segmentation of the observed images, which serves as the ground truth for segmentation. Then, we calculate the cross-entropy loss between the rendered semantics and the ground truth. Additionally, we construct a semantic-based road surface mask $M$ to filter out irrelevant information such as pedestrians and vehicles. The loss formulas for color and semantics are as follows:
\begin{equation}
L_{rgb} = \frac{1}{|M|} \sum M |C -C_{gt}|,
\end{equation}
\begin{equation}
L_{sem} = \frac{1}{|M|} \sum M\cdot CE(S, S_{gt}),
\end{equation}
where $C_{gt}$ represents the RGB ground truth and $S_{gt}$ represents the semantic ground truth. $CE()$ denotes the cross-entropy loss.

To better optimize the road surface elevation, we employ Lidar points for elevation supervision. Specifically, we query Lidar points within a certain neighborhood range of ground points, and their elevations are used as the ground truth $z_{gt}$ for the road surface elevation. The formula for elevation supervision loss is as follows:
\begin{equation}
L_z = \frac{1}{|M|} \sum M  |z - z_{gt}|.
\end{equation}

Furthermore, based on the smooth of the ground, we add a Laplacian smooth loss:
\begin{equation}
L_{smooth} = \sum_{i=1}^{N} \sum\limits_{j \in N(i)} \left| z_i - z_j \right|^2,
\end{equation}
where $N$ represents the number of vertices in the mesh, $N_i$ represents the set of vertices adjacent to the vertex $i$, and $h_i$ and $h_j$ represent the elevation values of the vertex $i$ and $j$, respectively. For each vertex $i$, calculate the sum of the squares of its elevation difference from its neighbor $j$, and then sum all the vertices. This encourages the model to learn to generate smoother mesh elevation predictions.

The total loss function is the combination of the above four losses.
\begin{equation}
L_{total} = \lambda_{rgb} L_{rgb} + \lambda_{sem} L_{sem} + \lambda_{z}L_z + \lambda_{smooth} L_{smooth},
\end{equation}
where, $\lambda_{rgb}$, $\lambda_{sem}$, $\lambda_{z}$, and $\lambda_{smooth}$ are the corresponding loss weights.

\noindent\textbf{Parameters optimization and the final outputs.} During the reconstruction process, the learnable parameters include the parameters of the elevation residual MLP, multiple RGB MLPs, and semantic and implicit color encoding. Once the training is completed, we predict the elevation of the road surface and decode the optimized implicit color feature using the trained color MLPs, resulting in a visual representation of the road surface that incorporates colors from different cameras and semantics.
\begin{figure*}[t] 
\center{\includegraphics[width=1.0\textwidth]{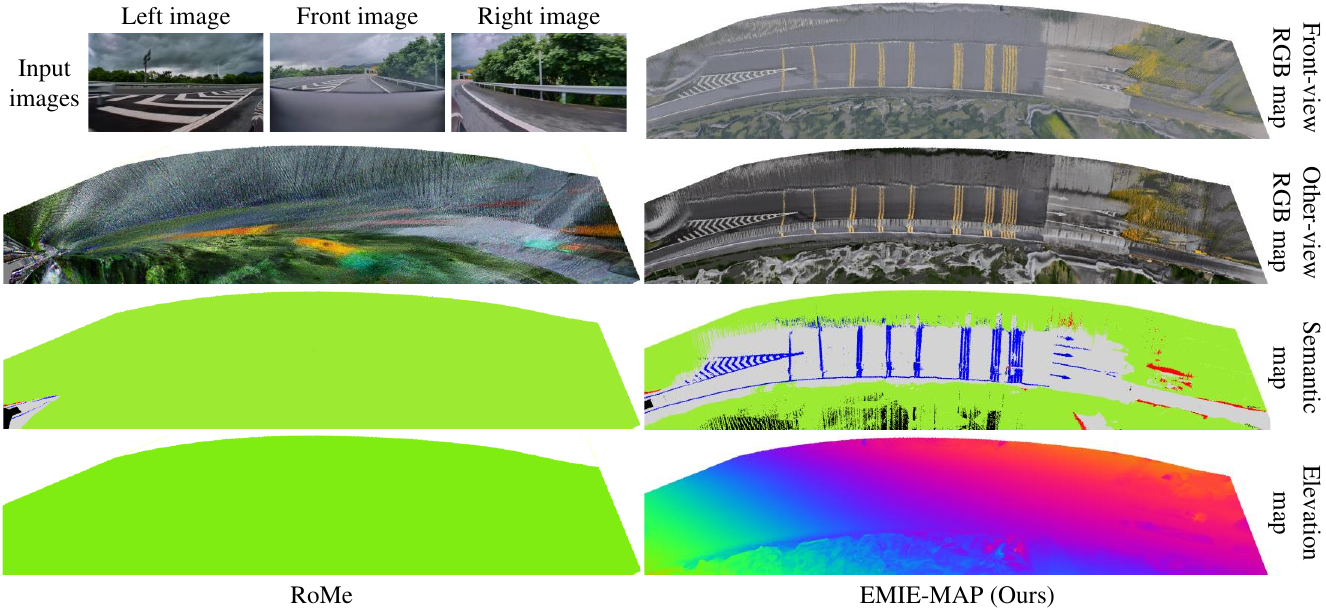}} 
\caption{Road surface reconstruction results in ramp scenes. Our method remains remarkable performance in ramp scenes, capturing road surface color, semantics, and elevation.}
\label{ramp}
\end{figure*}

\section{Experiment}
\label{sec:experiment}
\subsection{Experiment Setup}

\noindent\textbf{Datasets.}

We conducted experiments on the KITTI~\cite{geiger2013vision} dataset and a challenging custom dataset that includes diverse road surface scenes such as city streets, highways, slopes, tunnels, nighttime environments, and so on. Our data collection vehicle is equipped with a surround-view system consisting of seven cameras. Among them, there are two front-facing cameras, one wide-angle camera, and one telephoto camera, all with a resolution of $3840\times2160$. The other five cameras have the same model and a resolution of $1920\times1080$. They are positioned on the front-left, front-right, back-left, back-right, and back of the vehicle. Additionally, there is a 128-line Lidar sensor. The image capture frequency is 30Hz, while the Lidar operates at a frequency of 10Hz. 
Each data packet is approximately 30 seconds.

\noindent\textbf{Implementation details.}
The parameters to be optimized include the parameters of the elevation residual network and color decoders, as well as the semantic and color feature. We use the Adam~\cite{kingma2014adam} optimizer to optimize these parameters. The learning rate for the elevation residual MLP is set to 0.01, while the learning rate for the color MLPs is set to 0.005. The semantic learning rate is set to 0.1, and the learning rate for the color feature is set to 0.005. The loss weights are set as $\lambda_{rgb} = 1.0$,$\lambda_{sem} = 1.0$,$\lambda_{z} = 1.0$, and $\lambda_{smooth} = 1.0$.

 The road mesh resolution $a$ is set to 0.1m. The dimension of the color feature is 32. The elevation residual MLP consists of eight layers with a width of 128. The color MLPs consist of two layers with a width of 16. For each scene, a total of 5 epochs are trained with a batch size of 8. Due to the sufficient coverage of the road surface by the front-facing wide-angle camera and the front-left and front-right cameras, we only utilize images from these three perspectives for the reconstruction. All experiments are conducted on a server equipped with an NVIDIA A100 GPU. For more implementation details, please refer to the supplementary materials.

\noindent\textbf{Metrics.}

For road surface color and semantics, we project the reconstructed road surface onto the perspective of each camera to obtain rendered images. We evaluate the results using the Peak Signal-to-Noise Ratio (PSNR) for color fidelity and mean Intersection over Union (mIoU) for semantic segmentation accuracy. For road surface elevation, we evaluate the performance by calculating the average distance between the Lidar ground points and the reconstructed road surface, defined as Elev-error.

\noindent\textbf{Baseline.} We compare our method with RoMe~\cite{mei2023rome}, which is a road surface reconstruction method based on explicit mesh.

\subsection{Experimental Results}
For the custom dataset, we selected three challenging scenes for presenting the experimental results: city street, night, and ramp. For additional experimental results on more scenes, please refer to the supplementary materials. Tab. \ref{result} presents the road surface reconstruction evaluation results for the three scenes. Compared to RoMe~\cite{mei2023rome}, our method exhibits higher reconstruction accuracy and robustness in terms of road surface color, semantics, and elevation. Our method maintains remarkable reconstruction performance even under extreme lighting conditions and sharp elevation change, whereas RoMe~\cite{mei2023rome} fails completely.

\begin{figure}[t] 
\center{\includegraphics[width=1.0\columnwidth]{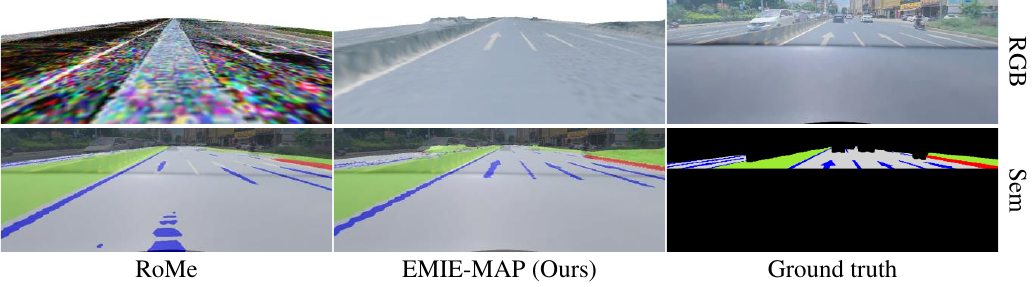}} 
\caption{Visualization of the rendered RGB and semantic images. It can be observed that the RGB map rendered by RoMe \cite{mei2023rome} appears chaotic, while our method is capable of rendering road surface colors accurately. Our method also achieves more accurate semantic reconstruction, as indicated by the higher consistency between the rendered semantic image and the ground truth RGB image overlay.}
\label{render}
\end{figure}

\begin{table*}[h]
\centering
\resizebox{\textwidth}{!}{
\begin{tabular}{c|ccc|ccc|ccc}
\toprule
\multirow{2}{*}{Scenes}  & &City street& & & Night & & & Ramp & \\
    &  PSNR $\uparrow$ &  mIoU (\%)  $\uparrow$ &  Elev-error (cm) $\downarrow$ &  PSNR $\uparrow$ &  mIoU (\%) $\uparrow$ &  Elev-error (cm) $\downarrow $ & PSNR $\uparrow$ &  mIoU (\%) $\uparrow$ &  Elev-error (cm) $\downarrow$  \\   
\midrule
RoMe ~\cite{mei2023rome}  & 17.31 & 94.87 & 45.08
                         & 0.99& 4.91& 331.23
                        &3.93 &3.95 & 496.24 \\
                        
EMIE-MAP w/o Lidar GT   & 24.44 & 95.12 & 17.08
                         & 22.12 & 94.14 & 27.34
                        & 19.20 & 85.07 & 93.12 \\
                        
EMIE-MAP w/  Lidar GT
& \textbf{26.75} & \textbf{95.27} &\textbf{2.35} 
&\textbf{24.53} &\textbf{96.02} & \textbf{3.57}
 &\textbf{20.74} &\textbf{88.89} &\textbf{4.33} \\

\bottomrule
\end{tabular}
}
\caption{Road surface reconstruction evaluation results. We conducted evaluations in three challenging scenarios. For color and semantic information, we evaluate the performance using the Peak Signal-to-Noise Ratio (PSNR) and mean Intersection over Union (mIoU) calculated from rendered RGB and semantic images. For the elevation information, we measure the average distance between the Lidar road points and the reconstructed road surface, defined as the Elev-error. Compared to RoMe ~\cite{mei2023rome}, our method demonstrates higher reconstruction accuracy and robustness. In the Night and Ramp scenarios, RoMe completely fails. The middle row shows the results without Lidar supervision, which are still significantly better than RoMe~\cite{mei2023rome}.}
\label{result}
\end{table*}

\noindent\textbf{City street scene.} 
Fig.~\ref{city} presents the results of road surface reconstruction in a city street scene. The input image reveals inconsistent luminance levels between the front view and other views. Our method is capable of reconstructing RGB maps that correspond to the camera's luminance levels, with the front view appearing brighter and the other view appearing darker. It is important to emphasize that both the front-view RGB map and other-view RGB map are decoded from the same optimized implicit color feature map. Different MLPs are employed to decode the RGB features into their corresponding colors. The RGB features capture camera-independent scene essential information, while the different MLPs learn camera-specific attributes. Furthermore, despite the presence of heavy traffic on the street, our method is still able to reconstruct clear and complete road surfaces and lane markings. This is attributed to the consistency of scenes from different viewpoints, where occlusions can be mutually supplemented.

Fig. \ref{render} presents the visualization of rendered RGB and semantic images. It can be observed that the RGB image rendered by RoMe \cite{mei2023rome} appears disordered due to the failure of optimization caused by inconsistent luminance between different cameras. In contrast, our method accurately renders road surface colors. From the overlay of the rendered semantic image and the ground truth RGB image, it is evident that our rendered semantic image exhibits higher consistency. This highlights the accuracy of our reconstruction method.

\noindent\textbf{Night scene.} 
Fig. \ref{night} illustrates the road surface reconstruction results in a night scene. The scene is extremely dark, with only a small bright area visible in the image. In such circumstances, our method can only reconstruct RGB maps that include the visible portion of the road surface. However, our method is still able to achieve excellent results in terms of semantic mapping, which highlights the versatility of our method.

\begin{figure*}[t] 
\center{\includegraphics[width=1.0\textwidth]{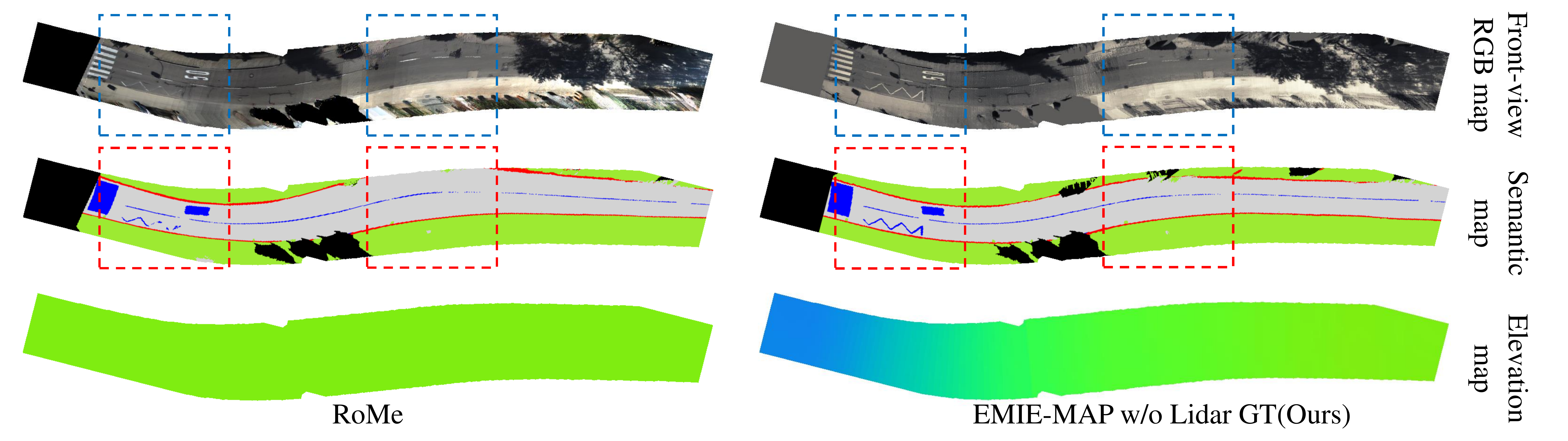}} 
\caption{Road surface reconstruction results in the KITTI-odometry dataset (sequence-09). EMIE-MAP is capable of reconstructing more accurate RGB, semantic, and elevation information than RoMe~\cite{mei2023rome} in curved and uphill scenes.}
\label{kitti}
\end{figure*}

\begin{wraptable}{l}{6cm}
    \centering
    \resizebox{\linewidth}{!}{
\begin{tabular}{l|ccc}
\toprule
Experiments   &  PSNR $\uparrow$ &  mIoU (\%) $\uparrow$ &  Elev-error (cm) $\downarrow$ \\
\midrule
RoMe~\cite{mei2023rome} & 20.87 & 94.57 & 107.98 \\
EMIE-MAP w/o Lidar GT &\textbf{21.93} & \textbf{94.76} & \textbf{28.12} \\
\bottomrule
\end{tabular}
}
\caption{Quantitative results in the KITTI-odometry dataset (sequence-09).}
\label{tab_kitti}
\end{wraptable}

\noindent\textbf{Ramp scene.} 
Fig. \ref{ramp} presents the road surface reconstruction results in a ramp scene. In this scene, there is a steep downhill slope on the road. Thanks to the designed trajectory-based elevation initialization and elevation residual prediction, our method accurately reconstructs the road surface elevation, thereby optimizing the scene's color and semantic information. It is important to note that if the road surface elevation optimization is inaccurate, there will be significant deviations in the RGB and semantic information based on projection optimization.

\noindent\textbf{KITTI dataset.}
To demonstrate the robustness and accuracy of EMIE-MAP, frames 1140 to 1237 of sequence-09 were selected. In this test segment, the vehicle traversed a section of road that involved both an incline and a curve. Similar to RoMe~\cite{mei2023rome}, we utilize monocular images from KITTI’s left RGB camera. To ensure a fair comparison with RoMe, we do not employ Lidar point cloud supervision for ground height. From Fig. \ref{kitti}, it can be seen that RoMe exhibits inconsistent road widths due to not considering changes in ground height. EMIE-MAP, even without Lidar point cloud supervision, can accurately model ground height and therefore achieve higher scores as shown in Tab~\ref{tab_kitti}.

\subsection{Ablation Study}

\begin{figure}[t]
\includegraphics[width=1.0\columnwidth]{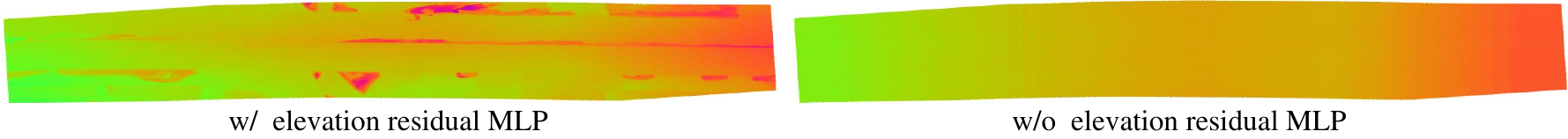}
\caption{Visualization of elevation maps in ablation experiments of elevation residual MLP. It is evident that using the elevation residual MLP results in more accurate elevation reconstruction.}
\label{ablation_z_fig}
\end{figure}

\noindent \textbf{Effect of elevation residual MLP.} Tab. \ref{ablation} (a) demonstrates the effect of elevation residual MLP
removal. Removing the elevation residual MLP results in a significant increase in elevation error. Fig. \ref{ablation_z_fig} illustrates that with the elevation residual MLP, the road surface elevation can be reconstructed more accurately with finer details.

\begin{figure}[t] 
\center{\includegraphics[width=1.0\columnwidth]{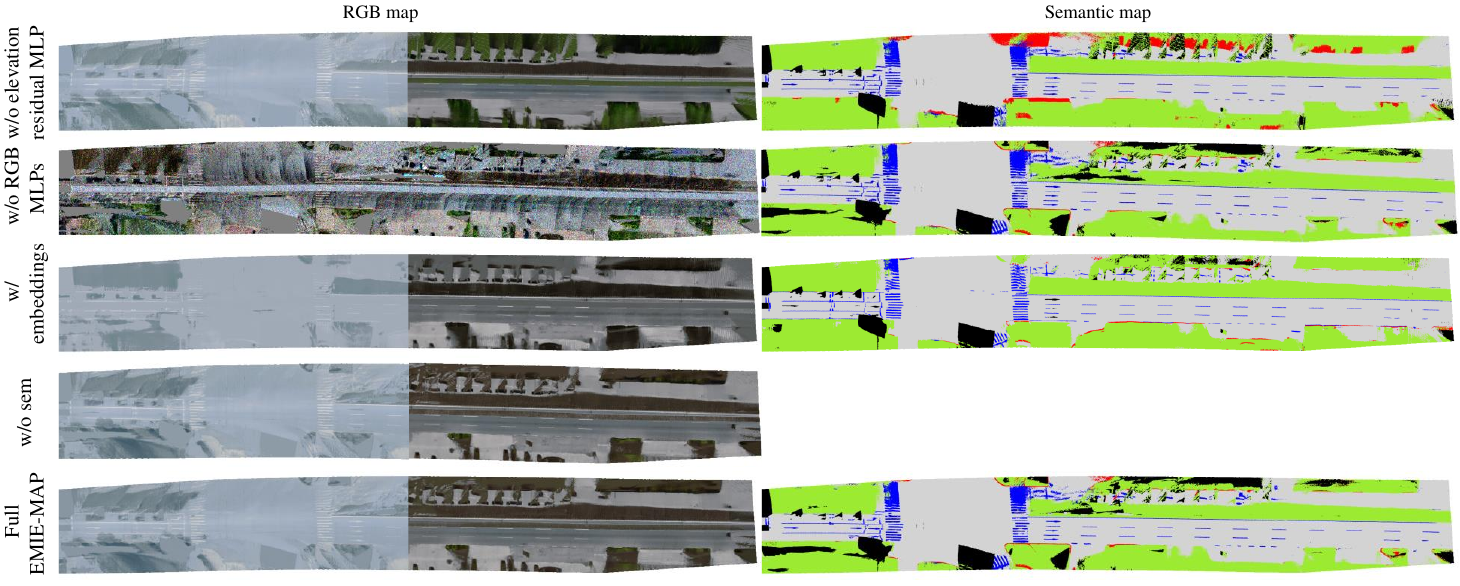}} 
\caption{Visualization of road surface reconstruction in ablation experiments. It is evident that using the elevation residual MLP results in more accurate elevation reconstruction. The inconsistent luminance between different cameras can cause explicit RGB optimization methods to fail. However, our color representation approach based on implicit color encoding and multiple RGB MLPs can address this challenge and achieve separate modeling for each camera.}
\label{ablation_fig}
\end{figure}

\noindent \textbf{Effect of the color representation.}
Tab. \ref{ablation} (b) showcases the effectiveness of the proposed color representation. After removing multiple RGB MLPs and directly optimizing the color parameters, the PSNR significantly declines. It can be observed from Fig. \ref{ablation_fig} that the inconsistent luminance between different cameras leads to the failure of direct color optimization. However, our designed representation, which utilizes consistent implicit color encoding and multiple RGB MLPs, addresses this challenge and enables the mapping of observed colors from different cameras. Using rendering-related embedding can also help to deal with this.
The results in Tab. \ref{ablation} (c) demonstrate that different MLPs is superior to a single MLP with different embeddings, as the modeling capacity of MLPs is stronger.

\begin{table}[h]
\centering
\resizebox{0.8\columnwidth}{!}{
\begin{tabular}{l|ccc}
\toprule
Experiments   &  PSNR $\uparrow$ &  mIoU (\%) $\uparrow$ &  Elev-error (cm) $\downarrow$ \\
\midrule
a. Ours w/o elevation residual MLP &26.63&94.35& 12.78 \\
b. Ours w/o RGB MLPs & 15.60 &95.16 & 2.49 \\
c. Ours w/ embedding &25.83 & 95.06 & 2.49  \\
d. Ours w/o sem &26.37 & - & 2.48  \\
e. Full EMIE-MAP (Ours) &\textbf{26.75} & \textbf{95.27} & \textbf{2.35} \\
\bottomrule
\end{tabular}
}
\caption{Ablation study of our design choices on the city street. The results validate the effectiveness of each of our innovations.}
\label{ablation}
\end{table}

\noindent \textbf{Effect of semantic information.} Tab. \ref{ablation} (d) demonstrates the results lacking semantic information. The introduction of semantics can enhance the color and geometric reconstruction results, as the semantic consistency across multiple perspectives can facilitate reconstruction optimization.

\section{Conclusion}
\label{sec:conclusion}
We propose EMIE-MAP, a novel large-scale road surface reconstruction method that combines explicit mesh and implicit encoding. We introduce an elevation optimization approach based on trajectory-based initialization and elevation residual prediction, which enables high-precision elevation reconstruction. Additionally, we propose a color representation method based on implicit RGB encoding and multiple RGB MLPs, allowing for distinct modeling of scene characteristics and camera-specific characteristics. This addresses the challenge of inconsistent camera luminance across different camera models. Experimental results in various challenging scenarios demonstrate that EMIE-MAP achieves high accuracy and robustness in road surface reconstruction.

\noindent \textbf{Limitations and future work.} EMIE-MAP can only handle single-data road surface reconstruction, we will consider the aggregation of multiple data in future work. Furthermore, EMIE-MAP relies on accurate camera poses. Further research is needed on how to perform road surface reconstruction in cases where camera poses are inaccurate.

\bibliographystyle{unsrt}  
\bibliography{references} 

\appendix

\newpage
\begin{center}
\noindent {\large  Appendix}   
\end{center}

\section{Overview}
\label{sec: Overview}
In this document, we present more details and several
extra results. In Sec.~\ref{sec: implementation}, we elaborate on the implementation details of our method. In Sec.~\ref{sec: more-result}, we present the road surface reconstruction results for additional scenarios.

\section{Further Implementation Details}
\label{sec: implementation}
In this section, we provide additional detailed implementation information.

\subsection{Hyperparameters}

The resolution of the road surface mesh is set to 0.1m, with a mesh range extending 15m on each side of the vehicle trajectory. Each vertex stores coordinates $p$, semantics $sem$, and implicit color encoding $l_c$. The coordinates are in three dimensions $(x, y, z)$, semantics include five categories: lane marking, curb, manhole, road, and background. The color encoding has a dimension of 32. $(x,y)$ is encoded into a 22-dimensional coordinate vector. The elevation residual MLP consists of eight layers with a width of 128, using the ReLU activation function. The color MLPs consist of two layers with width of 16, and with the ReLU activation function in the intermediate layer and the Sigmoid activation function in the output layer.

 The parameters to be optimized include the parameters of the elevation residual MLP and color MLPs, as well as the semantic and color encoding. We use the Adam~\cite{kingma2014adam} optimizer to optimize these parameters. The learning rate for the elevation residual MLP is set to 0.01, while the learning rate for the color MLPs is set to 0.005. The semantic learning rate is set to 0.1, and the learning rate for the color encoding is set to 0.005. The loss weights are set as $\lambda_{rgb} = 1.0$, $\lambda_{sem} = 1.0$, $\lambda_{z} = 1.0$, and $\lambda_{smooth} = 1.0$. For each scene, a total of 5 epochs are trained with a batch size of 8. 

 Due to the sufficient coverage of the road surface by the front-facing wide-angle camera and the front-left and front-right cameras, we only utilize images from these three perspectives for the reconstruction. All experiments are conducted on a server equipped with an NVIDIA A100 GPU.

\subsection{Evaluation Metrics}

For road surface color and semantics, we project the reconstructed road surface onto the perspective of each camera to obtain rendered images. We evaluate the results using the Peak Signal-to-Noise Ratio (PSNR) for color fidelity and mean Intersection over Union (mIoU) for semantic segmentation accuracy. For road surface elevation, we evaluate the performance by calculating the average distance between the Lidar ground points and the reconstructed road surface, defined as Elev-error.

\begin{figure*}[h] 
\center{\includegraphics[width=1.0\textwidth]{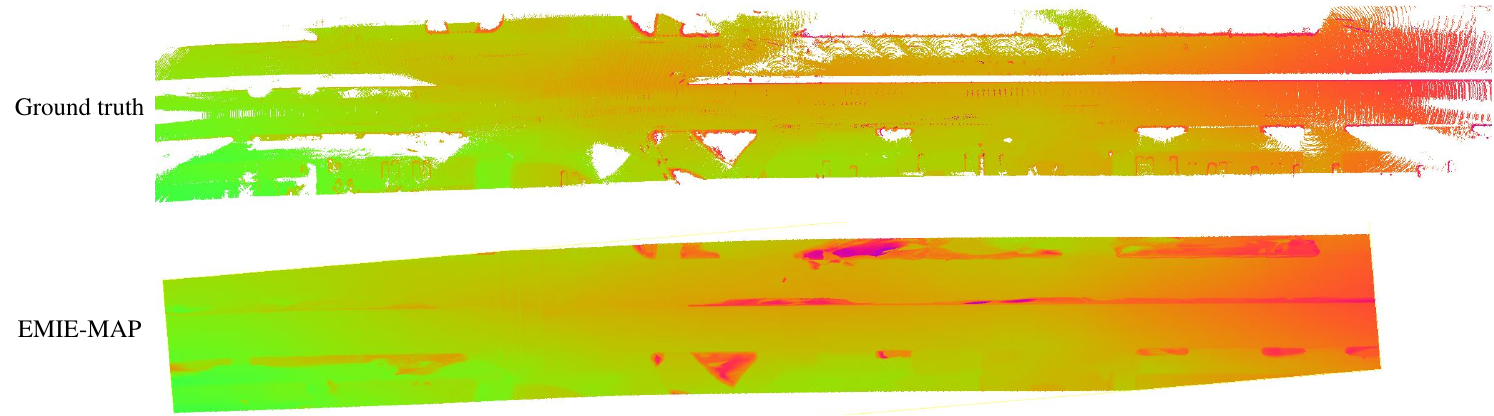} 
\caption{Elevation Evaluation. The top image shows the ground truth of the road surface obtained by concatenating the LiDAR point clouds, while the bottom image displays the road elevation map reconstructed using EMIE-MAP. For LiDAR points within the reconstruction area, we calculated the average distance from the LiDAR points to the reconstructed road surface as Elev-error.
}
\label{hight}
}
\end{figure*}

We performed object detection on each frame of the LiDAR point cloud to filter out vehicles and pedestrians on the road surface. Then, we concatenated all the LiDAR point clouds together based on vehicle trajectories to generate a local LiDAR map as the ground truth for the road surface. Fig \ref{hight} shows the ground truth of the road surface in the city street scene and the elevation map reconstructed using EMIE-MAP. For LiDAR points within the reconstruction area, we calculated the average distance from the LiDAR points to the reconstructed road surface as Elev-error.

\section{Per-Scene Road Surface Reconstruction Results}
\label{sec: more-result}

In this section, we provide road surface reconstruction results for more diverse scenarios, including highways, tunnels, curve, early morning and nighttime scenes. The dataset is sourced from our data collection vehicles. The public dataset used in the RoMe consists of flat scenes, with consistent color observation from different cameras. However, our surround-view cameras vary in model, resulting in inconsistent colors across different views, significantly increasing the challenge of this task. The results demonstrate the versatility of our algorithm, which achieves good reconstruction performance in various complex environments. Our method is capable of reconstructing RGB maps obtained from different camera models and semantic maps, and it accurately captures road surface elements such as lane lines and arrows. From the elevation maps, it is evident that our road surface reconstruction closely matches the height information from the LiDAR map. Our method not only effectively models changes in slope along the road but also captures height variations in details such as curbs. Compared to LiDAR maps, our reconstruction results provide a dense road surface mesh with color and semantic information, making them more suitable for downstream tasks such as BEV perception and data annotation.

\begin{figure*}[h] 
\center{\includegraphics[width=1.0\textwidth]{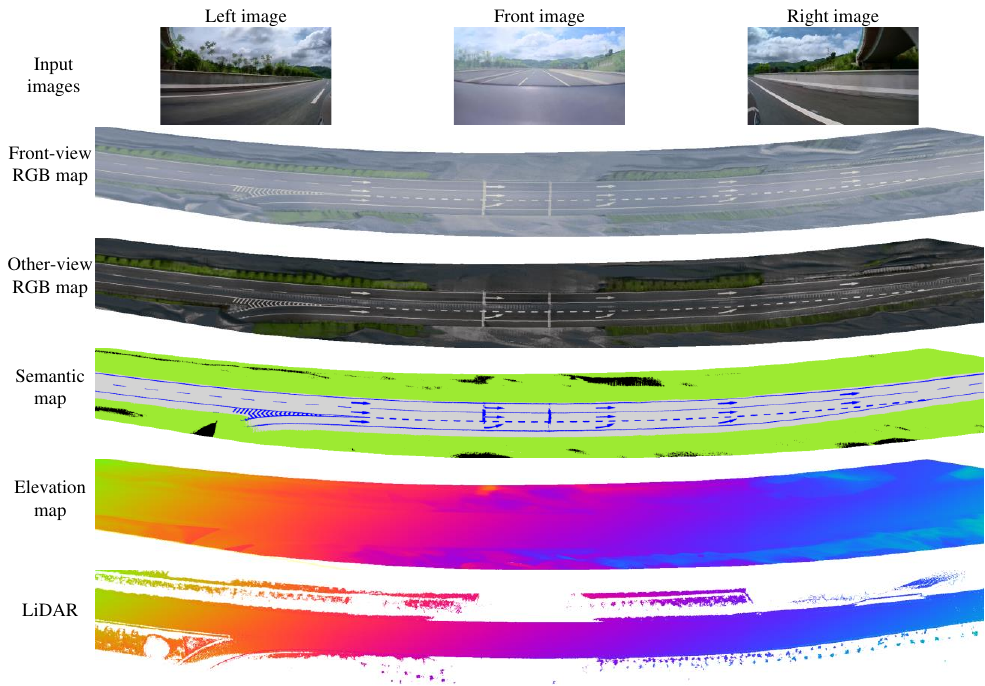} 
\caption{Road surface reconstruction results of EMIE-MAP in a highway scene.
}
\label{hight}
}
\end{figure*}

\begin{figure*}[h] 
\center{\includegraphics[width=1.0\textwidth]{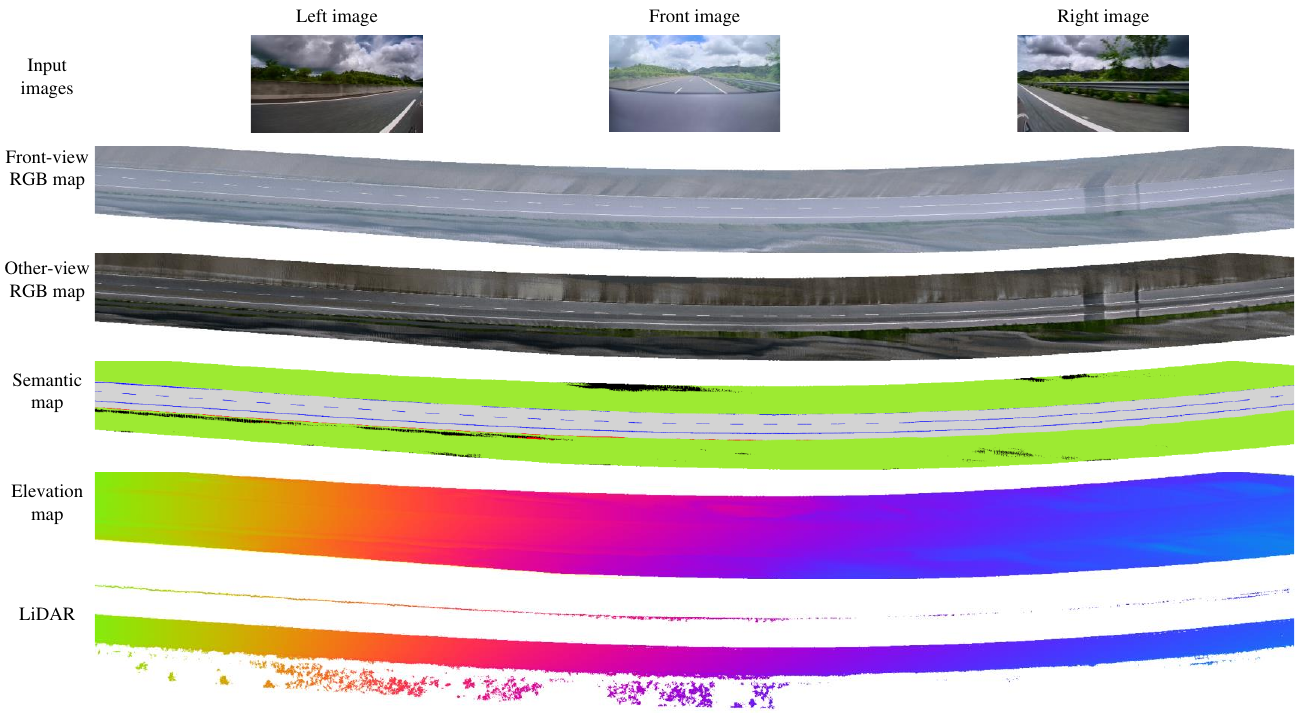} 
\caption{Road surface reconstruction results of EMIE-MAP in a highway scene.
}
\label{hight}
}
\end{figure*}

\begin{figure*}[h] 
\center{\includegraphics[width=1.0\textwidth]{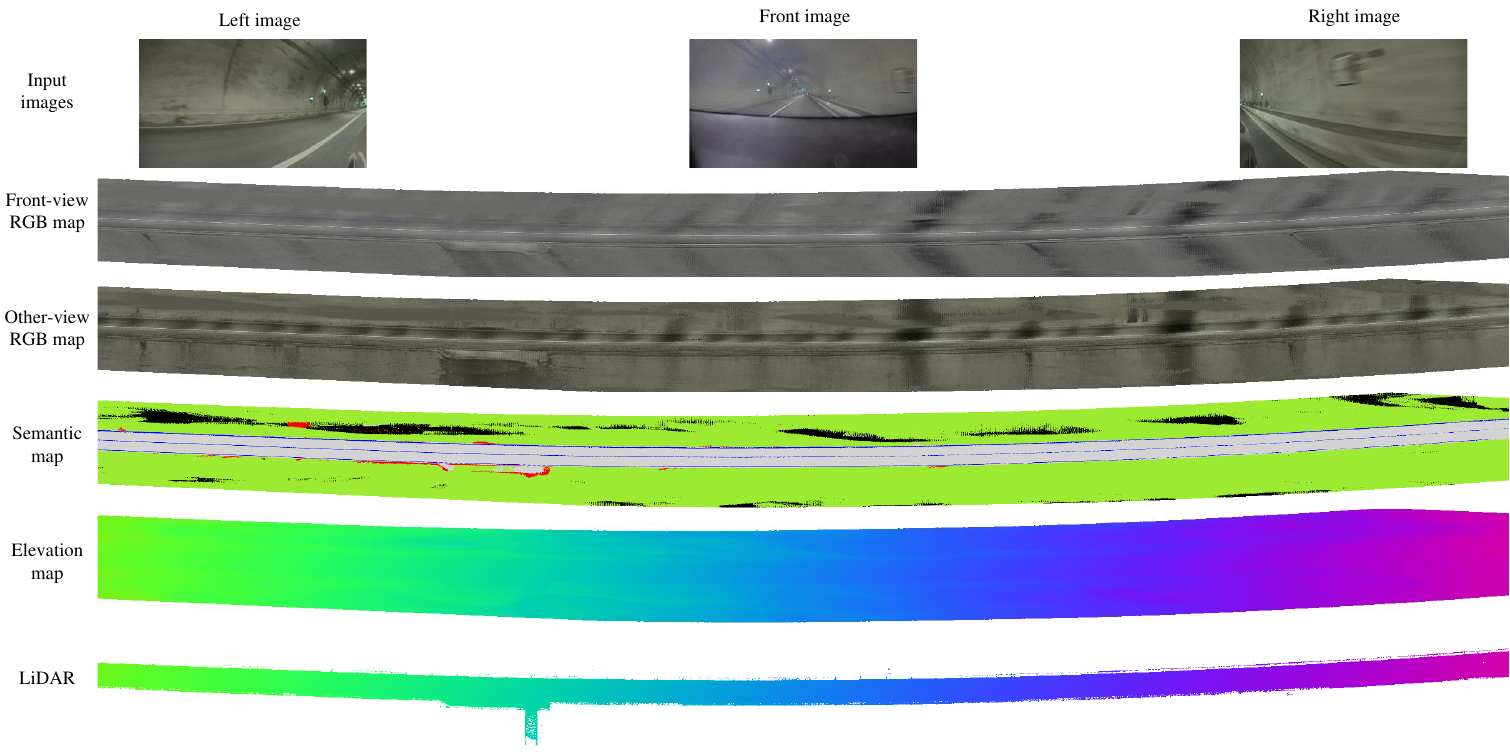} 
\caption{Road surface reconstruction results of EMIE-MAP in a tunnel scene.
}
\label{hight}
}
\end{figure*}

\begin{figure*}[h] 
\center{\includegraphics[width=1.0\textwidth]{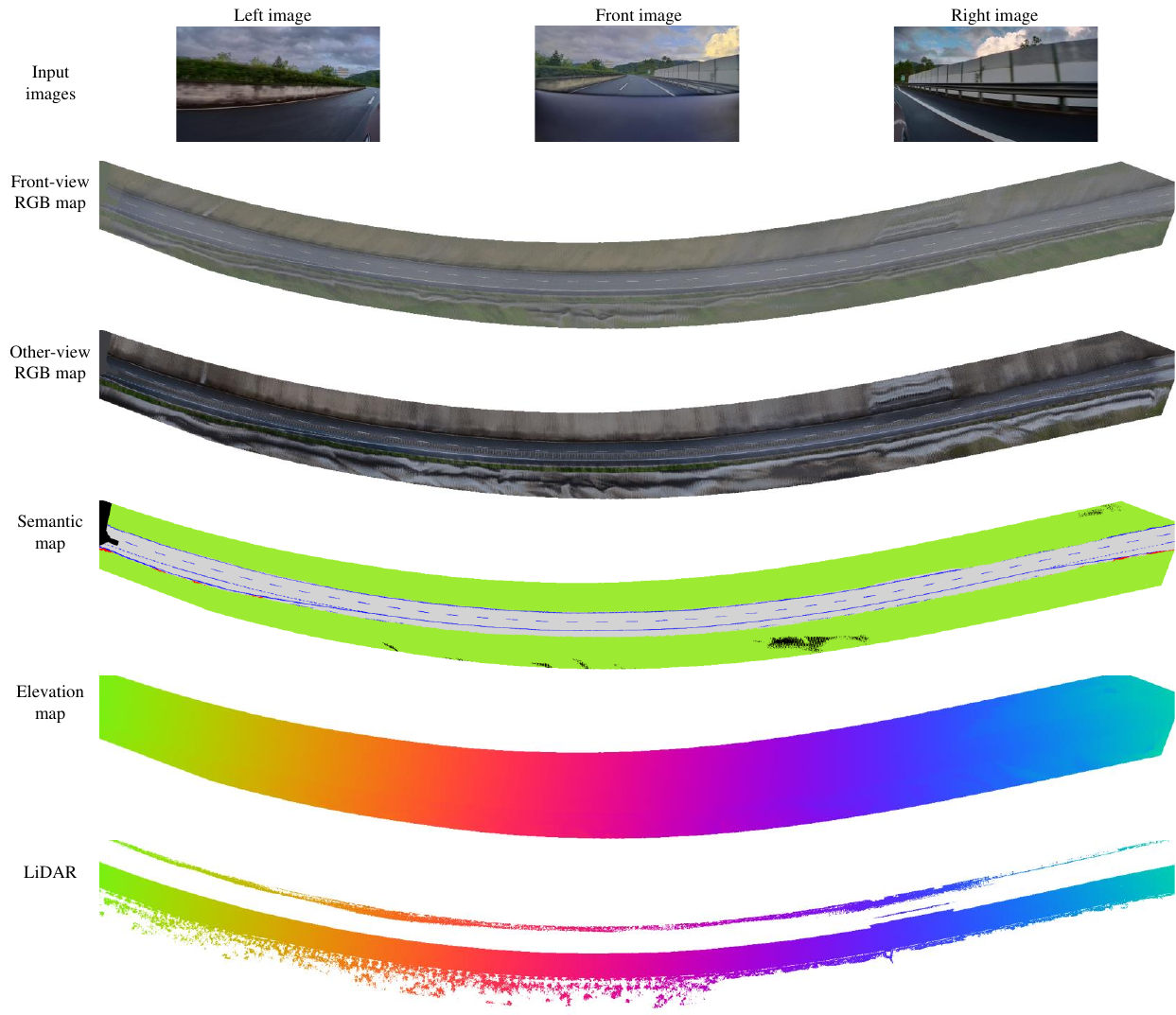} 
\caption{Road surface reconstruction results of EMIE-MAP in a curve scene.
}
\label{hight}
}
\end{figure*}

\begin{figure*}[h] 
\center{\includegraphics[width=1.0\textwidth]{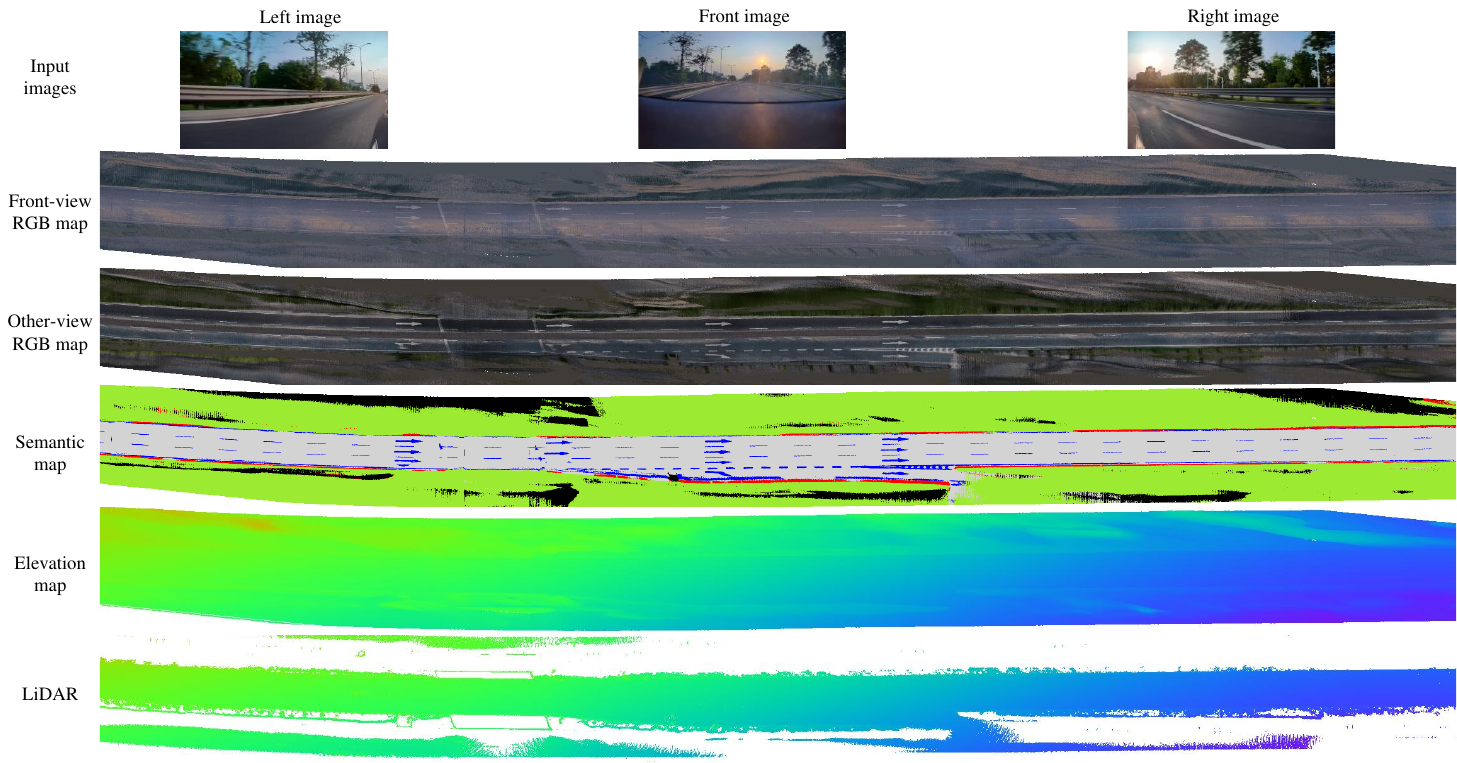} 
\caption{Road surface reconstruction results of EMIE-MAP in a early morning scene.
}
\label{hight}
}
\end{figure*}

\begin{figure*}[h] 
\center{\includegraphics[width=1.0\textwidth]{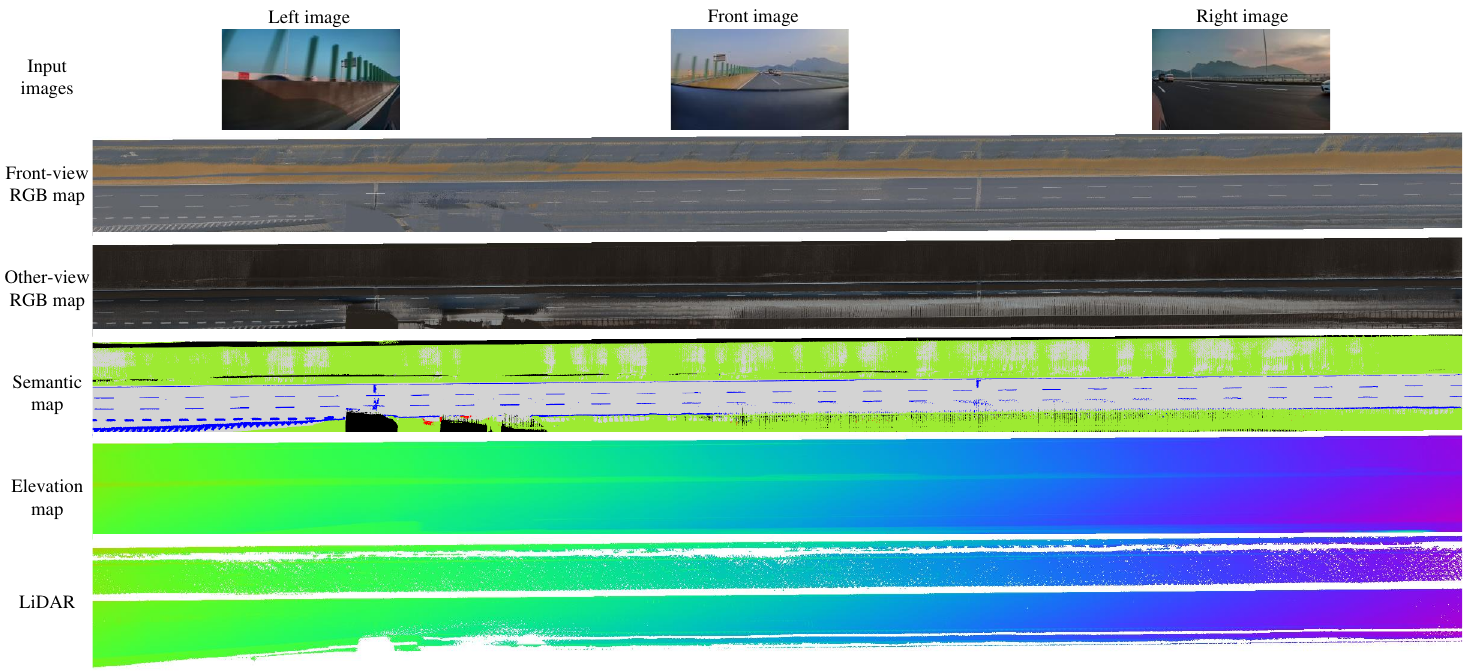} 
\caption{Road surface reconstruction results of EMIE-MAP in a early morning scene.
}
\label{hight}
}
\end{figure*}

\begin{figure*}[h] 
\center{\includegraphics[width=1.0\textwidth]{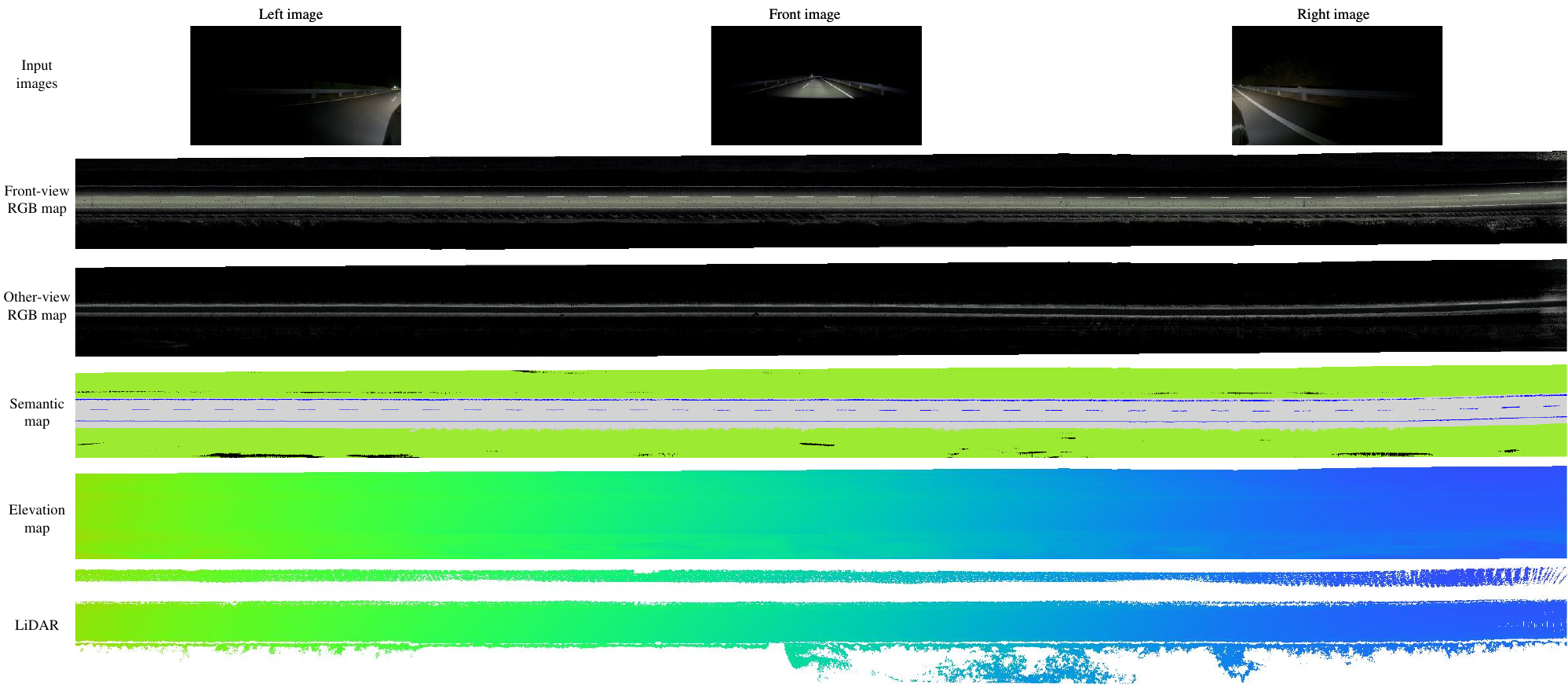} 
\caption{Road surface reconstruction results of EMIE-MAP in a night scene.
}
\label{hight}
}
\end{figure*}

\begin{figure*}[h] 
\center{\includegraphics[width=1.0\textwidth]{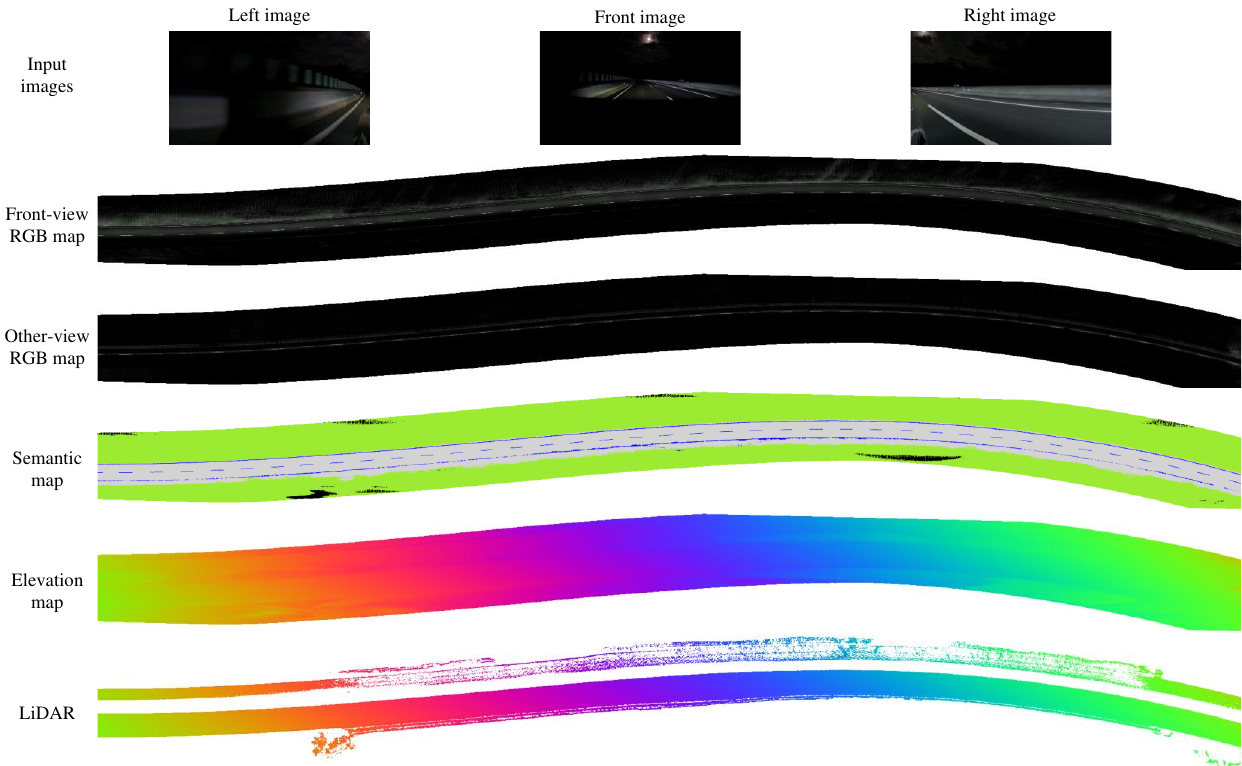} 
\caption{Road surface reconstruction results of EMIE-MAP in a night scene.
}
\label{hight}
}
\end{figure*}

\end{document}